\documentclass{article}
\usepackage{spconf,amsmath,graphicx}
\usepackage{amsfonts}
\usepackage{color}
\usepackage{comment}

\title{Adversarial Learning of Label Dependency: A Novel Framework  for Multi-class Classification}
\name{Che-Ping Tsai, Hung-Yi Lee}
\address{National Taiwan University\\
    \texttt{r06922039@ntu.edu.tw, tlkagkb93901106@gmail.com}}%
\begin{document}
\ninept
\maketitle
\begin{abstract}
Recent work has shown that exploiting relations between labels improves the performance of multi-label classification. 
We propose a novel framework based on generative adversarial networks (GANs) to model label dependency.  
The discriminator learns to model label dependency by discriminating real and generated label sets.
To fool the discriminator, the classifier, or generator, learns to generate label sets with dependencies close to real data. 
Extensive experiments and comparisons on two large-scale image classification benchmark datasets (MS-COCO and NUS-WIDE) show that the discriminator improves generalization ability for different kinds of models.
\end{abstract}

\begin{keywords}
Multi-label classification, Generative Adversarial Network 
\end{keywords}
\section{Introduction}
\label{sec:intro}
Multi-label classification is a fundamental but challenging problem in machine learning with applications such as multi-object recognition~\cite{kang2016object,kang2017t}, image classification~\cite{boutell2004learning}, text categorization~\cite{yang2018sgm}, and music categorization~\cite{tzanetakis2002musical}. 
In contrast to single-label classification, multi-label predictors must not only relate labels with the corresponding instances, but also exploit dependencies between labels due to label co-occurrences. 
Take for instance multi-label image categorization: \textit{beach} and \textit{sky} usually appear together in the same image, whereas \textit{airplane} and \textit{dog} do not often co-occur. 

A simple approach to using deep networks~-- for example CNNs~-- for multi-label classification is to recast the problem as multiple disjoint binary classification by replacing cross-entropy loss with logistic loss or ranking loss~\cite{gong2013deep}. To model label dependency, however, recent work has focused on capturing cross-label correlation using probabilistic graphical networks~\cite{li2016conditional,li2014multi}, dependency networks~\cite{guo2011multi}, recurrent neural networks (RNNs)~\cite{wang2016cnn}, and so on.

 In this work, we propose a new framework under which to train a multi-label classifier. 
 We use a generative adversarial network (GAN) to model the label distribution for multi-label classification.
 This framework is built upon a conditional GAN (cGAN). 
 The classifier here plays the role of a conditional generator, whose input is an instance, and which outputs a set of labels as with a typical multi-label classifier.  
A discriminator is trained to model label dependency: it takes an object and a set of labels as input, and outputs a score.
The set of labels comes either from the training data or from the output of the classifier, that is, the generator. 
The discriminator learns to discriminate the real and generated label sets.
To tell the real label from the generated ones, the discriminator must model the correlation of the input instances and their corresponding label sets as well as the label dependency of the label sets in the training data.
The classifier then learns to fool the discriminator by generating  label sets with what seems to be the correct dependencies, given an input instance by the discriminator. 
The classifier and discriminator are learned iteratively as in a typical GAN.

As the proposed framework is general and independent of the network architecture of the classifier, we believe the discriminator can be easily appended to other models to help learn label dependencies. 
Evaluation on two public multi-label image classification datasets shows that the discriminator facilitates generalization ability among CNNs with different architectures. 
To the best of our knowledge, this is the first attempt to utilize GANs for multi-label classification.

\section{Related work}
\label{sec:related work}

Multi-label classification has been widely studied in image classification.
A straightforward way to deal with multi-label classification is to decompose it into multiple binary classification tasks, such as binary relevance~\cite{tsoumakas2009correlation} using neural networks. To further improve performance, recent work has taken into account interdependency between labels.
Gong et al.~\cite{gong2013deep} evaluate various objectives with a CNN architecture, and find that weighted approximate ranking loss works best with CNNs. 
To better model the structure of label correlations, traditional graphical models have also been used for this task~\cite{li2016conditional,li2014multi}; latent space methods~\cite{yeh2017learning,bhatia2015sparse} have also been proposed.
Wang et al.~\cite{wang2016cnn} and Chen et al.~\cite{chen2017order} combine CNN and RNN to jointly embed images and semantic structure of labels in the same embedding space. 
Zhu et al.~\cite{zhu2017learning} further propose a spatial regularization network to capture both spatial and semantic relations. 

The work mentioned above mainly considers the global representation of the whole image, ignoring the relationships between semantic labels and local image regions, 
which is difficult to decipher given complex backgrounds.  
To handle such cases, Wei et al.~\cite{bhatia2015sparse} propose a Hypothesis-CNN-Pooling framework to aggregate the label scores of each proposal using category-wise max-pooling. Yang et al.~\cite{yang2016exploit} transform the multi-label recognition problem into a multi-class, multi-instance learning problem and make use of 
label-view information of the proposals  
to enhance features.
Newer work~\cite{chen2017recurrent,wang2017multi} uses long short term memory (LSTM) units to iteratively discover a sequence of attentional and informative regions and further predict labeling scores.

The proposed approach is independent of the above approaches. It is possible to further improve the above approaches with a discriminator. 

\begin{figure*}[!t]
   \centering
   \includegraphics[width=\linewidth]{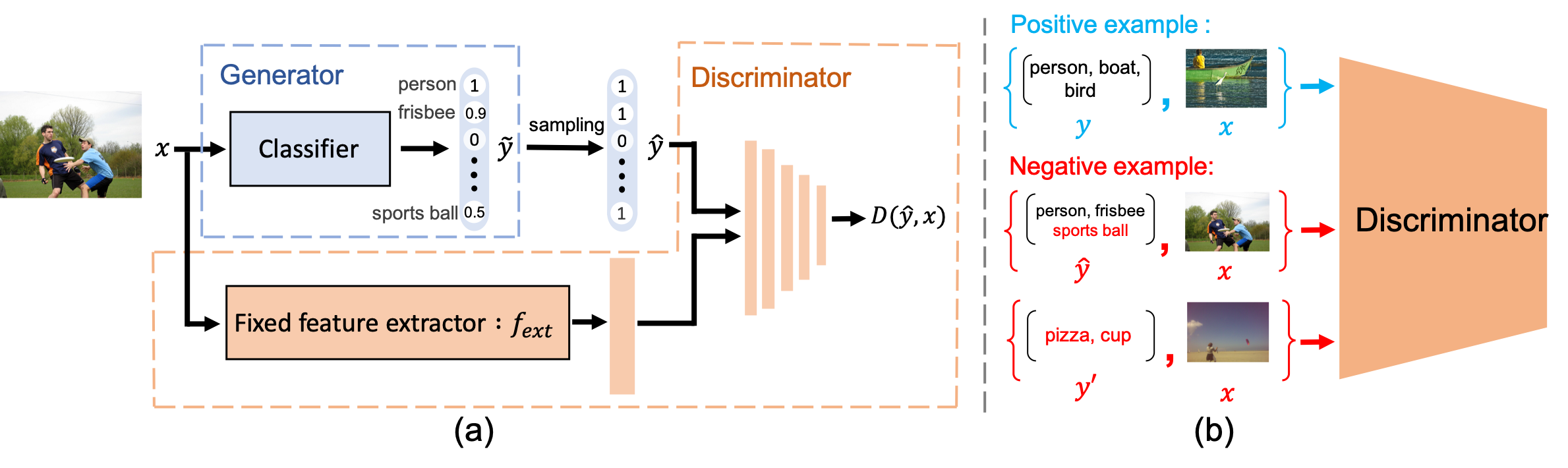}
   \caption{Proposed framework for multi-label classification, shown in (a). Fig (b) illustrates three kinds of discriminator inputs. The upper one is the positive example (\textit{real label set} $y$, \textit{image} $x$), a matched pair sampled from  real data distribution. The others are negative examples. The middle is the generated pair (\textit{generated label set} $\hat{y}$, \textit{image} $x$), where $\hat{y}$ is sampled from the output of generator $\tilde{y}$. The bottom is the mismatched pair (\textit{random sampled label set} $y^\prime$, \textit{image} $x$). The discriminator learns to assign low scores to the negative pairs, and 
   high scores to the positive pairs as genuine.}  % AMH: check
   \label{fig:framework}
\end{figure*}

%這段是多餘的
\begin{comment}
\subsection{Generative Adversarial Network}
Generative adversarial networks (GAN) are a leaning framework that consists of the generator the discriminator. The former tried to transform some known distribution (e.g. gaussian) to the distribution of real data, while the later measure the difference between a target and generated distribution. However, it required the composition of the generator and discriminator to be fully differentiable. In this paper, since the label distributions are discrete multi-hot vector, instead of using non-differential parts such as \textit{argmax} , we borrow the idea of Wasserstein GAN (WGAN)\cite{gulrajani2017improved}, which use the earth mover's distance on GAN so that it can measure the distance between a discrete and a continuous distribution. Moreover, inspired by Jang \textit{et al.}\cite{jang2016categorical} and Li \textit{et al.}\cite{li2018towards}, we also used gumbel sigmoid to reparameterize the sampling procedure to make it differentiable.
\end{comment}

\section{Methodology}
\label{sec:method}

The overview of the proposed framework is shown in Fig.~\ref{fig:framework}(a). 
Here, multi-label image classification is considered an instance of multi-class classification. 
Let $x$ denote an input image for which its corresponding ground-truth labels are $y \in \{0,1\}^{|S|}$, where $| S |$ is the number of labels (or  classes), and $y$ is a set of labels. 
The generator and discriminator in Fig.~\ref{fig:framework} are trained iteratively. 
That is, we fix one and update the other several times. 

\subsection{Classifier (Generator) and Discriminator}

The generator $G$\footnote{In contrast to the typical cGAN setting, the conditional generator here is conditioned on the image, not the class information.} is a classifier with sigmoid activation functions at the output layer. 
The classifier here can have a wide variety of architectures, for example, VGG-16~\cite{simonyan2014very}, Inception\_v3~\cite{szegedy2016rethinking}, Resnet-101~\cite{he2016deep}, or Resnet-152~\cite{he2016deep}. 
$G$ takes the input image $x$ and predicts the probability of each label to generate a probability distribution $\tilde{y}$. 
During the training phase, the predicted label set $\hat{y}$ is sampled from $\tilde{y}$ by considering the sigmoid output as a probability.
During testing, labels whose sigmoid outputs are larger than 0.5 are in the output label set.

The discriminator $D$ receives a label set $y$ (or $\hat{y}$) and an image $x$, and produces a score, $D(y,x)$ (or $D(\hat{y},x)$), which represents the measurement of how ``real'' the label distribution is and the degree of match between the image and the label set. 
$D$ contains a feature extractor network $f_{\mathit{ext}}$ which produces $z=f_{\mathit{ext}}(x)$ as 
the feature vector for image $x$.  % AMH: check
Then a feedforward network takes $z$ and $y$ (or $\hat{y}$) as input and outputs a single value $D(y,x)$ (or $D(\hat{y},x)$).
During the testing phase, only the generator is used; the discriminator is left unused. 

\subsection{Classifier Training}

$G$ is pretrained as a typical multi-class classifier
by minimizing the binary cross-entropy loss $\mathcal{L}_{\mathit{logistic}}$ using the ground truth labels $y=[y_1,y_2,...,y_{|S|}]^T$ and the predicted probability of each label $\tilde{y} =[\tilde{y}_{1},\tilde{y}_{2},...,\tilde{y}_{|S|}]^T$:
\begin{equation}
  \mathcal{L}_{\mathit{logistic}} = \mathbb{E}_{(x,y)\sim{\mathit{data}}} [ \mathop{\sum_{i=1}^{|S|}} y_i log\,\tilde{y}_{i}  + (1-y_i)log(1-\tilde{y}_{i}) ],
  \label{eqn:logistic_loss} 
\end{equation}
where $(x,y)$ is 
% a pair of image and label set sampled from training data,
an $\langle$image, label set$\rangle$ pair sampled from training data,  
and $\tilde{y}$ is the output label distribution of $G$ given $x$ sampled from the training data.
As all labels are considered independently in $\mathcal{L}_{\mathit{logistic}}$, 
the classifier learned by minimizing   $\mathcal{L}_{\mathit{logistic}}$ is not guaranteed to model dependencies between labels.

During the iterative training, the loss function of generator $G$ is 
\begin{equation}
  \mathcal{L}^\prime_{G} = \mathcal{L}_{G} + \alpha \mathcal{L}_{\mathit{logistic}}, 
  \label{eqn:generator_loss} 
\end{equation}
where $\alpha$ determines the scale of the logistic loss and
\begin{equation}
  \mathcal{L}_{G} = - \mathbb{E}_{x\sim{\mathit{data}},\hat{y} \sim G(x)}[D(\hat{y},x)], 
  \label{eqn:generator_dis_loss} 
\end{equation}
where $x$ is an input image sampled from training data.
$\hat{y}$ is the output label set of $G$ sampled from $\tilde{y}$. 
$D(\hat{y},x)$  is the score assigned by the discriminator given an $\langle$image, label set$\rangle$ pair. 
With (\ref{eqn:generator_dis_loss}), the generator learns not only to minimize the logistic loss $\mathcal{L}_{\mathit{logistic}}$, but also  to produce a reasonable combination of labels to fool the discriminator $D$ by maximizing the value $D(\hat{y},x)$.
Because $D$ takes the whole label set $y$ or $\hat{y}$ as input, it uses the dependency between the labels to discriminate real and generated label sets.
Therefore, the $G$ learned from $D$ takes into account label dependency.

Equation~(\ref{eqn:generator_dis_loss}) requires the composition of the generator and discriminator to be fully differentiable. 
Here, since the label set $\hat{y}$ is a discrete multi-hot vector, we use the Gumbel-softmax trick for a Bernoulli distribution of each label~\cite{jang2016categorical,li2018towards}, which we here term \textit{Gumbel sigmoid}, to reparameterize the sampling procedure to make it differentiable.

\subsection{Discriminator Training}
To train the discriminator $D$, (\textit{real label set} $y$, \textit{image} $x$)  sampled from real data distribution serves as positive examples.
For negative examples, we not only provide $D$ with the generated pairs, (\textit{generated label set}, $\hat{y}$, \textit{image} $x$), but also the mismatched pairs, (\textit{random sampled} $y^\prime$, \textit{image} $x$), where $y^\prime$ is a label set randomly sampled from the training data which does not correspond to image $x$~\cite{reed2016generative}. Fig.~\ref{fig:framework}(b) illustrates these three kinds of discriminator inputs.
Here, we use a Wasserstein GAN with a gradient penalty (WGAN-gp).
The loss function of discriminator $\mathcal{L}_{D}$ is
\begin{equation}
 \begin{split}
&  \mathcal{L}_{D} =   - \mathbb{E}_{(x,y) \sim \mathit{data}} [D(y,x)]    \\
&   + \frac{1}{2}\mathbb{E}_{x \sim \mathit{data}, \hat{y} \sim G(x)}[D(\hat{y},x)] +  \frac{1}{2} \mathbb{E}_{x \sim \mathit{data}, y^\prime \sim \mathit{data}} [ D(y^\prime,x) ] \\
&  + \lambda \mathcal{L}_{gp}.
  \label{eqn:discriminator_loss} 
 \end{split}
\end{equation}
In the first term, an image $x$ and its label set $y$ is sampled from the training dataset, from which $D$ learns 
to assign large values to genuine samples.   % AMH: check
In the second term, the image $x$ and the label set $\hat{y}$ predicted by the generator is generated, 
for which $D$ is expected to assign a small value.  % AMH: check
To help the discriminator to learn how to discriminate mismatched image/label pairs, we use negative sampling as (the third term):
we randomly select mismatched $\langle$image $x$, label set $y^\prime\rangle$ pairs from the training data, to which $D$ also learns to assign low scores.  
That is, the discriminator learns to assign low scores to two kinds of errors: the generated label set from the classifier and realistic label sets with the wrong images that mismatch the condition information. 

The last term $\mathcal{L}_{gp}$ is the gradient penalty~\cite{gulrajani2017improved}; $\lambda$ determines the gradient penalty scale.
\begin{equation}
\mathcal{L}_{gp} = \mathbb{E}_{x\sim{\mathit{data}}, \hat{y} \sim G(x), y^* \sim\mathit{interp}(\hat{y},y^*) } [ (\|\nabla D(y^*,x)\|-1)^2 ] 
  \label{eqn:gp} 
\end{equation}
We interpolate the real label $y$ and the generated label $\hat{y}$ with random weights between 0 and 1 to generate $y^*$, and apply the gradient penalty to $y^*$. 
This gradient penalty is essential to stabilize the training.

\section{Experiment}
We evaluated the proposed approach on the MS-COCO benchmark with 80 labels, and NUS-WIDE with 81 labels. As performance measures we used macro/micro precision, macro/micro recall, and macro/micro F1-measure. Note that macro/micro P/R/F1 scores are abbreviated as O/C-P/R/F1, respectively. Generally, O/C-F1 is more important \cite{sorower2010literature}. 

In the experiments, we compared the proposed approach with 
WARP~\cite{gong2013deep}, CNN-RNN~\cite{wang2017multi}, order-free CNN-RNN with visual attention (Att-RNN)~\cite{chen2017order}, and RLSD~\cite{multzhang2018i}, which also models label dependency. 
To show that the proposed approach is general, we compared four kinds of generator architectures: CNNs, VGG-16~\cite{simonyan2014very}, Inception\_v3~\cite{szegedy2016rethinking}, Resnet-101~\cite{he2016deep}, and Resnet-152~\cite{he2016deep}. 

\label{sec:exp}
\subsection{Implementation details}

%這部分後面再講清楚
%The feature extractor network $f_{ext}$ shares the same architecture as the generator $G$ except the final prediction layer. We extract visual features from \textit{global pooling} layer, which is just in front of the final prediction layer. It remains fixed after pretraining to increase the stability of training cGAN.
The generator was pretrained on ImageNet with 1000 categories. %($|S|=1000$)  as a typical multi-class classifier.
We removed the output layer of the pretrained generator as the feature extractor network $f_{\mathit{ext}}$ in the discriminator.
Its parameters remained fixed during training.
For the discriminator, the image features $z=f_{\mathit{ext}}(x)$ and label set were both linearly projected onto 256-dimension vectors, and then simply concatenated and fed into 8 fully connected layers of 512 dimensions with leaky relu activation functions. 
It is common to update the generator and discriminator with different numbers of steps. According to our experimental observation, as the pretrained generators were strong enough, we trained the discriminators 3 times per generator update. For each discriminator training iteration, we randomly sampled 3 batches for 3 kinds of inputs: matched pairs, generated pairs, and mismatched pairs, as indicated in Fig.~\ref{fig:framework}(b); this is essential for training stability.
% and performing label generation is rather easier than image generation in normal setting of GAN

We based the whole optimization process on the Adam optimizer with a learning rate of 0.0001 for both the generator and the discriminator. 
The logistic loss weight of the generator $\alpha$ and the gradient penalty scale of the discriminator $\lambda$ were both set to 10. In Gumbel sigmoid, the inverse temperature was set to 0.9 without annealing.

We followed \cite{wang2015towards} for our data augmentation strategies. Specifically, we first resized the image to $256 \times 256$, after which we extracted five patches (four corner patches and the center patch) with a size from the set \{256,224,192,168,128\}. Finally, we resized the patches to $224 \times 224$.  
For testing, we simply resized all images to $224 \times 224$ and conducted single-crop evaluation. 
%In addition, we abandon the discriminator and use the output probabilities of generator as the final prediction. If the estimated probabilities for any label are greater than 0.5, the labels are predicted as positive. -> 前面說過了

\subsection{Experimental results}

\subsubsection{Microsoft COCO}

Microsoft COCO (MS-COCO) is a large-scale dataset for object detection, segmentation, and image captioning. It has also been used for multi-label classification. It comprises a training set of 82,081 images, and a validation set of 40,137 images from 80 classes. Since the ground truth labels of the 2014 challenge are not available, we followed \cite{chen2017order} and \cite{wang2017multi} in utilizing the validation set to evaluate our methods. 

From Table~\ref{table:coco_comparison}, we see that the baselines of Inception\_v3, Resnet-101, and Resnet-152 without the discriminator outperform other methods in C-F1 and O-F1 due to the advanced deep neural network structures.
Moreover, all four models trained with WGAN-gp achieve higher C/O-F1 scores than the baselines without discriminator, which further suggests that modeling label correlation improves multi-label classification. 
The performance gain is less obvious when the baseline model is stronger. 
For example, there is a 4.4\%/2.7\% performance gain in C/O-F1 for VGG-16 but only 0.6\%/0\% for Resnet-152. 
For deeper networks such as Resnet-101 and Resnet-152, they may implicitly learn label dependencies due the huge number of hidden layers; this limits the usefulness of WGAN-gp.
We also note that models which use WGAN-gp achieve higher recall but lower precision. 
We find that baseline models predict 2.09 labels per instance on average, whereas models which use WGAN-gp predict 2.61 labels, which is about 25\% higher than the previous and results in higher F1 scores.

\begin{table}[htp]
\centering
\begin{tabular}{|c||ccc|ccc|}
\hline
\centering Methods  & C-P & C-R & C-F1 & O-P & O-R & O-F1  \\
\hline
\centering  WARP  & 59.3 & 52.5 & 55.7 & 59.8 & 61.4 & 60.7   \\
\centering  CNN-RNN & 66.0 & 55.6 & 60.4 & 69.2 & 66.4 & 67.8 \\
\centering Att-RNN & 71.6 & 54.8 & 62.1 & 74.2 & 62.2 & 67.7 \\
\centering RLSD & 67.6 & 57.2 & 62.0 & 70.1 & 63.4 & 66.5 \\
\hline
\hline
\centering  VGG-16  & 74.2 &  44.8 & 56.0 & 77.6 & 52.5 & 62.6 \\ 
\centering + WGAN-gp & 62.6 & 58.3 & 60.4 & 67.5 & 63.3 & 65.3 \\
\centering  Inception\_v3  & 76.4 &  52.8 & 62.4 & 80.0 & 58.8 & 67.8 \\ 
\centering + WGAN-gp & 70.5 & 58.2 & 63.8 & 73.2 & 63.8 & 68.2 \\
\centering  Resnet-101  & 76.2 &  53.4 & 62.8 & \textbf{80.8} & 58.9 & 68.1 \\ 
\centering + WGAN-gp & 70.5 & \textbf{58.7} & \textbf{64.0} & 72.3 & \textbf{64.6} & 68.2 \\
\centering  Resnet-152  & \textbf{76.6} &  53.9 & 63.3 & 80.6 & 59.6 & \textbf{68.6} \\ 
\centering + WGAN-gp & 71.4 & 57.9 & 63.9 & 73.6 & 64.2 & \textbf{68.6} \\

\hline
\end{tabular}
\caption{Multi-label classification results on MS-COCO with 80 labels. Results of WARP, CNN-RNN, and RLSD are reported with the top 3 labels.}
\label{table:coco_comparison}
\end{table}

\begin{figure}[t!]
	\resizebox{\linewidth}{!}{
		\begin{tabular}{ccc}
		    &\includegraphics[height=2cm]{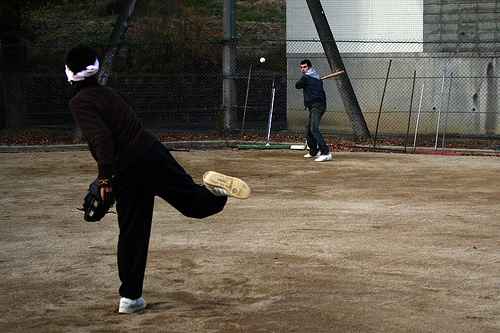} &
			\includegraphics[height=2cm]{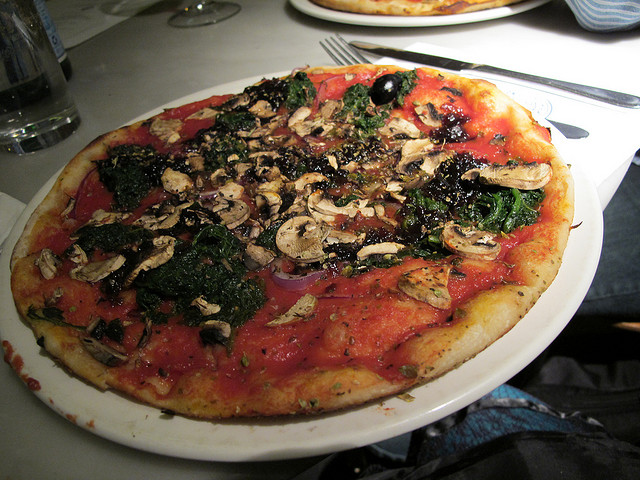} \\
			& (A) & (B)\\
            \hline
            \centering Ground truth:&
            \begin{tabular}{@{}c@{}}person, sports ball, \\ baseball bat, baseball glove\end{tabular}  &
            \begin{tabular}{@{}c@{}}wine glass, cup, fork,\\ knife, pizza, dining table \end{tabular} \\
            Resnet-101 &
            \color{red}{person, baseball bat}&
            \color{red}{fork, knife, pizza, dining table} \\
            \begin{tabular}{@{}c@{}}Resnet-101 +\\
             WGAN-gp \end{tabular} &
            \color{blue}{\begin{tabular}{@{}c@{}}person, sports ball,\\ baseball bat, baseball glove \end{tabular}} & \color{blue}{\begin{tabular}{@{}c@{}}wine glass, cup, fork,\\
            knife, pizza, dining table \end{tabular}}  \\
            \hline
             & & \\
             &\includegraphics[height=2cm]{./img/example3.jpg} &
			\includegraphics[height=2cm]{./img/example4.jpg} \\
			& (C) & (D)\\
            \hline
            \centering Ground truth:&
            chair, couch, bed, book  &
            person, laptop \\
            Resnet-101 &
            \color{red}{couch, tv, book}&
            \color{red}{person,laptop} \\
            \begin{tabular}{@{}c@{}}Resnet-101 +\\
             WGAN-gp \end{tabular} &
            \color{blue}{\begin{tabular}{@{}c@{}}chair, couch, tv,\\ laptop, book \end{tabular}} & \color{blue}{\begin{tabular}{@{}c@{}} laptop, mouse,\\
            keyboard \end{tabular}}  \\
            \hline
	\end{tabular}}
	\caption{Multi-label classification results from MS-COCO. With WGAN-gp, classifiers better predict smaller-sized image objects. For example, (A): Resnet-101 + WGAN-gp correctly predicts \textit{baseball glove} and \textit{sports ball} based on observations of \textit{person} and \textit{baseball bat}. However, in (D), it incorrectly relates \textit{mouse} and \textit{keyboard} to \textit{laptop}.} 
	\label{fig:results_examples} 
\end{figure}

This shows that with WGAN-gp, the classifier better models label dependencies and thus extracts more labels that are not detected by the original classifier. Examples of multi-label classification results are shown in Fig.~\ref{fig:results_examples}.

\subsubsection{NUS-WIDE}

NUS-WIDE is a web image dataset which contains 269,648 images and associated tags from Flickr. The images are further manually annotated into 81 concepts. Following the experimental settings of WARP~\cite{gong2013deep} and Att-RNN~\cite{chen2017order}, we removed images without annotations and used 150,000 images for training and 59,347 images for testing.
The results are reported in Table~\ref{table:nuswide_comparison}.
MS-COCO and NUS-WIDE show similar trends.

\begin{table}[htp]
\centering
\begin{tabular}{|c||ccc|ccc|}
\hline
\centering Methods  & C-P & C-R & C-F1 & O-P & O-R & O-F1  \\
\hline
\centering  WARP  & 31.7 & 35.6 & 33.5 & 48.6 & 60.5 & 53.9   \\
\centering  CNN-RNN & 40.5 & 30.4 & 34.7 & 49.9 & 61.7 & 55.2 \\
\centering Att-RNN & 59.4 & 50.7 & 54.7 & 69.0 & 71.4 & 70.2 \\
\centering RLSD & 44.4 & 49.6 & 46.9 & 54.4 & 67.6 & 60.3 \\
\hline
\hline
\centering  VGG-16  & 53.3 & 24.9 & 33.9 & 73.9 & 59.6  & 66.0 \\ 
\centering + WGAN-gp   & 51.6 & 34.3 & 41.2 & 68.8 & 67.3 & 68.1 \\
\centering  Inception\_v3 & 67.9 & 44.1 & 53.5 & 74.7 & 64.8 & 70.3 \\ 
\centering + WGAN-gp & 62.4 & 50.5 & \textbf{55.8} & 71.4 & 70.9 & \textbf{71.2} \\
\centering  Resnet-101 & 67.0 & 44.0 & 53.1  & 76.3  & 65.0 & 70.2 \\ 
\centering + WGAN-gp & 59.6 & \textbf{51.8} & 55.4 & 68.9 & \textbf{72.8} & 70.8\\
\centering  Resnet-152 & \textbf{69.1} & 41.8 & 52.1 & \textbf{75.9} & 65.1 & 70.1 \\ 
\centering + WGAN-gp  & 65.2 & 46.2 & 54.1 & 71.3 & 70.8 & 71.1 \\

\hline
\end{tabular}
\caption{Multi-label classification results on NUS-WIDE with 81 labels. Results of WARP, CNN-RNN, and RLSD are reported with the top 3 labels.}
\label{table:nuswide_comparison}
\end{table}

\subsubsection{Ablation study}
In this section, we show all the mechanisms described in Sect.~\ref{sec:method}. %including conditional discriminator, negative sampling, gumbel sigmoid and the loss of WGAN are useful for multi-label classification. 
Table~\ref{table:ablation} reports the macro/micro F1 scores of different type of models. In this experiment, we used Resnet-101 as the generator and performed classification on MS-COCO.

Rows (a) and row (b) show the results of Resnet-101 with all the mechanisms and the baseline model trained with logistic loss, as reported in Table~\ref{table:coco_comparison}. 
The classifier models in rows (c), (d), and (e) have the same network architecture as in rows (b), but we remove some GAN training tricks. 
In rows (c) and (d), we do not perform negative sampling. 
That is, the third term of $\mathcal{L}_{D}$ in Eq.~\ref{eqn:discriminator_loss} is removed, and the weight of the second term becomes 1.
In row (d), we replace the conditional discriminator with an unconditional one. 
Therefore, the only discriminator input is $y$ or $\hat{y}$. 
The discriminators here need only distinguish between real and generated label sets. 
The scores in row (c) and (d) are both less than the baseline. 
In row (e), as we directly feed the generator continuous output distribution $\tilde{y}$ to the discriminator, Gumbel sigmoid is not needed. %(after sigmoid activation)  
However, since the real data $y$ is discrete, the generator must sharpen the distribution $\tilde{y}$, which reduces performance.
%Thus, the generator may suffered from the same problem above. 
%In row (f), we use the objective function of the original GAN. 
%The results showed that WGAN-gp  performs better. 

\begin{table}[htp]
\centering
\begin{tabular}{|l||c|c|}
\hline
Methods  &  C-F1 & O-F1  \\
\hline
(a): Resnet-101 & 62.8 & 68.1 \\ 
(b): Resnet-101 + WGAN-gp & \textbf{64.0} &  \textbf{68.3} \\
\hline\hline
(c): (b) w/o negative sampling & 62.2 & 67.5\\
(d): (b) w/o conditional discriminator& 62.5& 67.6 \\
(e): (b) w/o Gumbel sigmoid & 62.3 & 67.1\\
%(f) Resnet-101 + GAN & 61.3 & 67.0\\
\hline
\end{tabular}
\caption{Macro/micro F1 scores with/without specific modules. Results are evaluated on MS-COCO with the Resnet-101 generator.}
\label{table:ablation}
\end{table}

\section{Conclusion}
\label{sec:conclusion}

In this paper, we propose a novel framework for multi-class classification.\footnote{This work was financially supported by the Ministry of Science and
Technology of Taiwan.}
Inspired by GAN, the discriminator learns to model label dependency by discriminating real and generated label sets.
To fool the discriminator, the classifier learns to generate label sets with dependencies close to real data. 
Extensive experiments and comparisons on two large-scale image classification benchmark datasets show that with this discriminator, F1 scores are improved across different classifier models. 
In future work, because the proposed idea is a general framework for multi-class classification, we will apply the proposed approach on multi-class classification tasks other than image classification.


\nocite{*}
\bibliographystyle{IEEEbib}
\bibliography{refs}

\end{document}